# Highly accurate digital traffic recording as a basis for future mobility research: Methods and concepts of the research project HDV-Mess

**Laurent Kloeker[1*], Fabian Thomsen[1], Lutz Eckstein[1], Philip Trettner[2], Tim Elsner[2], Julius Nehring-Wirxel[2], Kersten Schuster[2], Leif Kobbelt[2], Michael Hoesch[3]**
1. Institute for Automotive Engineering, RWTH Aachen University, Germany
2. Visual Computing Institute, RWTH Aachen University, Germany
3. Neurosoft GmbH, Germany
*laurent.kloeker@ika.rwth-aachen.de

**Abstract**
The research project HDV-Mess aims at a currently missing, but very crucial component for addressing important challenges in the field of connected and automated driving on public roads. The goal is to record traffic events at various relevant locations with high accuracy and to collect real traffic data as a basis for the development and validation of current and future sensor technologies as well as automated driving functions. For this purpose, it is necessary to develop a concept for a mobile modular system of measuring stations for highly accurate traffic data acquisition, which enables a temporary installation of a sensor and communication infrastructure at different locations. Within this paper, we first discuss the project goals before we present our traffic detection concept using mobile modular intelligent transport systems stations (ITS-Ss). We then explain the approaches for data processing of sensor raw data to refined trajectories, data communication, and data validation.

**Keywords:**
Digital traffic recording, mobile intelligent transport systems stations

**Introduction**

Traffic data collection is an important core aspect of the development and safeguarding of automated driving functions. Large data sets are needed to cover a wide range of situations and scenarios in real traffic. The collected data can be used to train and optimize algorithms for their intended use to guarantee the best possible level of safety. Naturalistic data from real traffic is of particular importance. The knowledge gained about human driving and movement behaviour in road traffic can later be transferred to simulation models and used for further research and development purposes. This is based on raw sensor data from real traffic. Conventional data collection of traffic data for automated driving functions is carried out with sensor vehicles equipped with a large number of sensors on the vehicle's exterior and recording the surrounding traffic situation during the test drive [1]. However, the disadvantage here is a limited field of vision and the occlusion of road users. For example, a truck in front of the sensor vehicle obscures the view of the traffic behind it and limits the range of information for this scene considerably. Furthermore, only the traffic events of a static map section around the sensor vehicle can be detected. The motion of the sensor vehicle itself results in an always dynamic map section. This prevents permanent and targeted recordings at highly interactive static road cross-sections, such as traffic intersections or freeway ramps. Another possibility for traffic data collection is the use of drones [2]. The advantage here is the avoidance of object occlusion due to the bird's eye view. During the data collection, however, constant supervision by a human is required and the recording times are limited by the battery capacities of the drones.

For this reason, the research project HDV-Mess, which is funded by the European Regional Development Fund (ERDF) and the German federal state of North Rhine-Westphalia, deals with the conceptual design and development of mobile and modular ITS-Ss for highly accurate traffic data collection. With the help of a suitable conceptual design and development of such ITS-Ss, measurements can be carried out fully self-sufficiently for several days on arbitrary road cross-sections of all domains, e.g. urban, rural, and freeway. Using several mobile ITS-Ss per measurement cross-section, for example, the installed sensor technology can record traffic events from multiple perspectives in parallel, thus guaranteeing full coverage. An elevated sensor position on the ITS-Ss also helps to capture traffic events from better angles and minimize object occlusion. Due to the modular design of the ITS-Ss, sensors can be easily exchanged and their quality compared and evaluated with regard to their possible applications. This enables further insights to be gained in the field of modern sensor technologies, which will remain

Highly accurate digital traffic recording as a basis for future mobility research: Methods and concepts of the research project HDV-Mess highly relevant for a long time to come due to their importance for automated driving.

**Related work**

Traffic recording with infrastructure sensors has already been tested several times in past research projects. However, the focus was almost exclusively on stationary ITS-Ss. Initial trials took place, for example, as part of the Ko-PER [3] project, in which a road intersection was equipped with stationary Light Detection and Ranging (LiDAR) sensors. A technical extension of such a use case was the research intersection of the Application Platform Intelligent Mobility (AIM) [4]. Here, stationary camera and radar sensors were installed to permanently record the traffic situation.
Currently, there are numerous other research activities related to the conceptual design and construction of large-scale test fields for traffic data collection. However, the focus here is also primarily on stationary ITS-Ss. The sensor technology used in all activities is based on a combination of optional camera, radar, and LiDAR sensors. Based on the research results of the AIM research intersection, the test field Lower Saxony was established, which covers not only the urban domain but also the domains of rural roads and highways [5]. The test fields of the Providentia++ [6] and Diginet-PS [7] research projects have also already been set up and are capable of recording real traffic events. Another large-scale test field, which is currently under construction, is being implemented in the ACCorD research project [8]. Numerous stationary ITS-Ss are also being used here on urban and rural test sites and the highway, using camera sensors and LiDARs to record the surrounding traffic situation.

**Project objective**

The aim of the research project HDV-Mess is the highly accurate recording of traffic events at various relevant locations and the collection of sensor and traffic data as a basis for the development and validation of current and future sensor technologies as well as automated driving functions. The recorded raw sensor data will first be stored on a central data server and then used to extract traffic trajectories. Thus, on the one hand, the raw sensor data will serve as a basis for the research fields of Big Data and machine learning in the vehicle sector. On the other hand, the collected trajectories from real traffic create a valuable naturalistic traffic data basis. The accuracy requirement for the trajectories to be extracted is a few centimetres. Besides, the module is to serve as a basis for the location-independent testing of intelligent transport systems in real traffic. To realize this project, a concept for a mobile modular fully autonomous kit of ITS-Ss for highly accurate traffic data acquisition will be developed. The conceived mobile ITS-Ss are designed in the form of trailers with a hydraulically extendable mast, on whose masthead the sensors are mounted. Also, the computing infrastructure and the power generator including intermediate storage batteries are installed inside the trailer. In total, four of these mobile ITS-Ss are to be constructed, which will be supplemented by a mobile receiver server to form a fully-fledged system. The mobile receiver server is used for bundled local data storage at the respective measurement cross-sections. To collect sufficiently diverse traffic data per measurement cross-section, it is necessary to permanently record it over a long period. The power generators and intermediate storage batteries integrated for full self-sufficiency enable a continuous recording period of several weeks without the need for an external power supply or human supervision. Thus, official approval for road traffic shall enable a temporary installation of a sensor infrastructure at different road cross-sections. Finally, the collected measurement data will be validated with the help of reference measurements by sensor vehicles or drones and evaluated with regard to their quality. This will provide an important basis for the research and development of innovative systems and functions.

**Traffic detection concept**

Fig. 1 shows our overall traffic detection concept. Four mobile ITS-Ss serve as key components recording the raw data and are constructed as follows: The basis of each ITS-S is a trailer equipped with a mast extendable up to a height of about 8 m, on which four sensors are mounted, as shown in Fig. 2.

Highly accurate digital traffic recording as a basis for future mobility research: Methods and concepts of the research project HDV-Mess

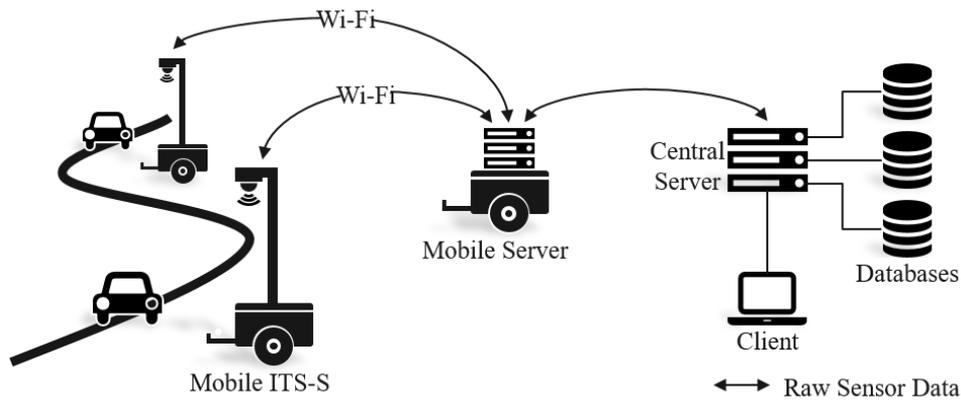

**Figure 1 - Overview of the traffic detection concept used in HDV-Mess.**

Two Ouster OS1-128 LiDAR sensors are placed on top of each other in the centre and can be adjusted in their pitch angles to always cover the complete width of an observed scene with a high resolution of laser reflections. On both sides of them, a FLEXIDOME IP starlight 8000i camera by Bosch is located recording videos at a 4K resolution. These cameras can be remotely controlled to turn in any direction at a large range of focal lengths allowing various use-cases. For instance, they can point downwards to supplement the LiDAR perception, to the side to enlarge the station's field of view (FOV), or, when multiple ITS-Ss are placed next to each other, the sensors of neighbouring stations can be oriented to see the same scene from different angles resulting in a better three-dimensional perception. The two types of sensors give complementary information about the perceived traffic with the cameras delivering clear colour information that can be useful e.g. for detecting lanes or classifying objects, whereas the LiDARs record their three-dimensional shape more precisely than cheaper radar sensors could do. Overall, each ITS-S produces four streams of raw data, which are sent to a mobile server via Wi-Fi. The raw data is anonymized, then stored on exchangeable hard drives. After having finished a recording at a specific site, these drives are mounted to a central server where the data is further processed to trajectories and stored in databases for later access. All these data processing steps are explained in detail in the following sections. An exemplary overview of the recorded raw sensor data and the output of object detection and classification algorithms is also shown in Fig. 2.

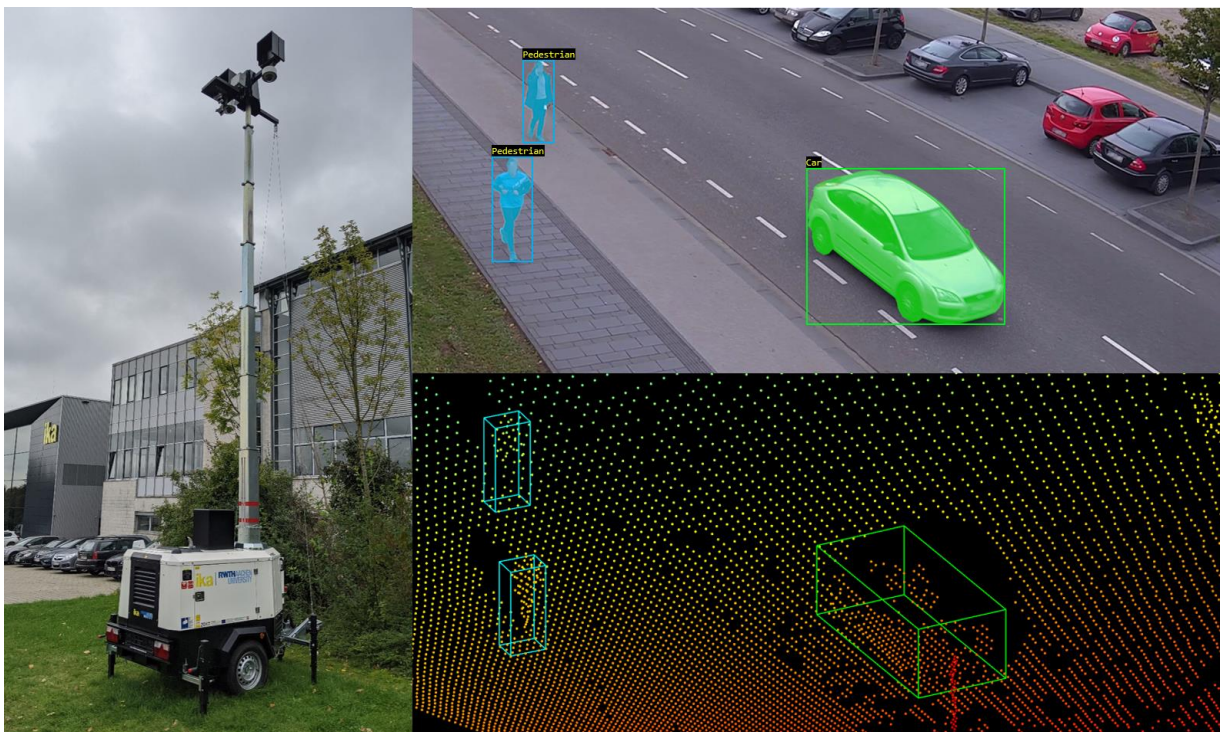

**Figure 2 - Mobile and modular ITS-S designed in HDV-Mess (left) and output of object detection and classification algorithms on recorded sensor raw data of a traffic scene (right).**

Highly accurate digital traffic recording as a basis for future mobility research: Methods and concepts of the research project HDV-Mess

**Data processing pipeline**

*Pipeline overview*

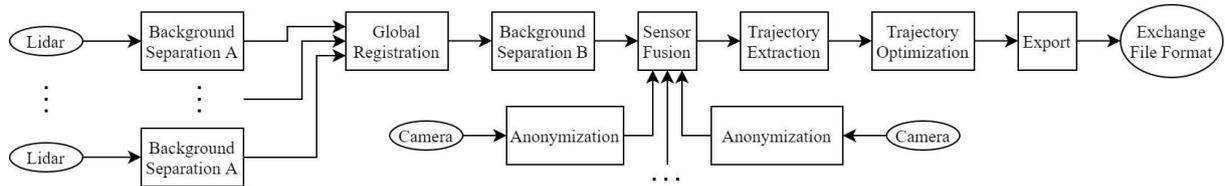

**Figure 3 - Overview of our data processing pipeline. Data preprocessing and sensor fusion is used to create a denoised and precise shared scene from which our highly accurate trajectories can be extracted.**

An overview of our data processing pipeline is shown in Figure 3. The system mainly uses the 3D data from the LiDAR sensors as this has proven most reliable. RGB camera data is fused in later in the process and is mainly used for metadata and additional classification.

Each LiDAR produces a stream of timestamped 3D points. In a first step, we consider each LiDAR separately and conservatively classify points that belong to the scene background, i.e., not to any vehicle or pedestrian, as "static", based on temporal statistics analysis. After this pre-classification, a global registration step is performed to transform all LiDAR points into a shared global coordinate system. With access to data from other LiDARs, a second background separation pass is performed to drastically reduce the number of points that were previously not classified as "static" but should have been. At this point, the quality of the combined point cloud is sufficient to project (properly anonymized) camera data onto the points. This marks the end of the preprocessing and sensor fusion.

The remaining steps of the pipeline are high-quality trajectory extraction and export. For that, we first perform a preliminary trajectory extraction that just separates different trajectories and collects all points belonging to the same trajectory. For each such extracted trajectory, we perform an optimization step that simultaneously computes a high-quality trajectory and a 3D model of the object. Finally, in an export step, we write out trajectories to many supported file formats. In the following, we describe each step in more detail.

*Background separation*

The background separation is done in two phases to solve a chicken-and-egg situation. Proper background separation needs data from all LiDARs but without a global registration, there is no shared coordinate system. However, global registration is unreliable without the background separation as non-static data causes false correspondences.

Thus, we first conservatively classify static points of each LiDAR separately. Our measuring stations are robust enough that there is only negligible per-station drift in the short term, thus we can assume that the timestamped per-LiDAR points are already in a shared coordinate system. The LiDAR scans can be thought of as a 2D grid of depth values where each value is updated with 10 Hz. Not all values are updated at the same time but in a rotating column-wise manner. Each grid entry corresponds to a single scan ray. For each scan ray, we consider all depths values measured along this ray. These are samples of a continuous distribution of depth values. Intuitively, this distribution is a superposition of the static scene and all dynamic objects that intersect with the scan ray. Over a longer period, the dynamic part will only amount to a small probability mass, e.g., below 15 % of all samples. The static part is often uni-modal around the "background depth", but can also be multi-modal, especially at depth discontinuities and challenging objects like trees. Given the samples of a scanning ray, we employ a mode detection algorithm to find all modes that amount to more than a given percentage of samples, e.g., all modes consisting of at least 20 % of the samples. Every mode corresponds to a 3D position. We use the point cloud consisting of all modes as the "static geometry" of the considered LiDAR. All points of a LiDAR that are within a certain threshold (e.g., 30 cm) of its static geometry are removed. The rest is considered non-static for now and contains dynamic objects as well as noise and areas with too few samples. However, this conservatively estimated static geometry is already sufficient for global registration.

Highly accurate digital traffic recording as a basis for future mobility research: Methods and concepts of the research project HDV-Mess

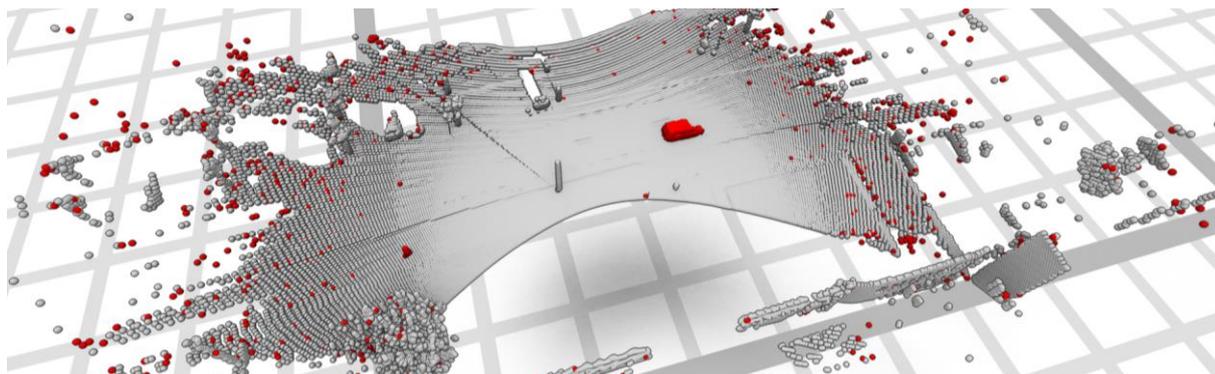

**Figure 4 - The single-LiDAR background separation is used to separate static (grey) from non-static (red) points with a low false-positive rate. The static part can then be used for global registration.**

The second phase of the background separation is applied after the registration. As we now have a global, shared coordinate system, we can use data from all LiDARs to improve the static/non-static classification. The static points of all LiDARs are merged into a single point cloud and organized in a bounding volume hierarchy for fast queries. Every non-static point that has a static point within a certain neighbourhood (e.g., 30 cm), is removed. This is similar to what is done in phase A, but this time contains data from all LiDARs and is thus able to remove most noise artefacts within our scene. Noise at the scene boundary can be removed with a similar heuristic. The result of this second phase is a quite clean separation of foreground and background, i.e., dynamic and static parts of the scene. Only a few clusters of points at the fringes of the scene are still misclassified. These are easily discarded during trajectory extraction.

*Global registration*

The point-cloud data produced by a single LiDAR is recorded in a local, LiDAR-centric, coordinate frame. To combine the data from multiple LiDARs into a single (global) point-cloud, the points from each LiDAR must be transformed (or registered) into a single global coordinate system.
We assume that a LiDAR's position in 3D space remains constant for the entire length of the measurement and at this point ignore potential dynamic movement caused by outer factors like wind or vibrations caused by passing vehicles. This assumption allows us to perform registration only once for an entire recording.
To perform the registration, we only rely on points that are classified as static and accumulate points over a few seconds to compensate for noisy data. We then select one of the accumulated input point-clouds as reference and use a probabilistic registration algorithm [9] to determine the transformation of the remaining point-clouds into the same coordinate system as our reference point-cloud. In some cases, it is necessary to roughly align the input point-clouds by hand to assure that the registration algorithm snaps to the correct optimum. After all point-clouds have been registered onto each other, we additionally determine the up-direction and apply a corresponding transformation to all point-clouds. This step is purely cosmetic and can be performed by estimating the dominant normal direction, which can be uniquely determined since large parts of the recorded data usually consist of the flat street surface.

*Sensor fusion*

At this point in the pipeline, all points are in a shared, global coordinate system and classified into static and non-static. We can now fuse in the anonymized camera data. All cameras are calibrated and their intrinsic and extrinsic parameters (relative to the LiDARs of the same station) are known. Thus, for each 3D point, we can compute its projection into each camera, discarding those that are not inside the camera's field of view and those that are occluded by other points, which can be tested via cone query. This allows us to annotate each 3D point with a set of colours and (u,v) coordinates for each successful projection.

Highly accurate digital traffic recording as a basis for future mobility research: Methods and concepts of the research project HDV-Mess

*Trajectory extraction*

The following steps of our pipeline are concerned with high-quality trajectory extraction. In this first step, our aim is to collect all non-static points that belong to the same traffic participant, i.e., to the same trajectory. The quality of the trajectory is irrelevant for this step, as long as the trajectories of different participants are not mixed. Thus, we can use a simple, yet effective grid-based scheme:

All non-static points are inserted into a virtual 4D grid based on their 3D position and timestamp. A sensible resolution of the spatial coordinates is around 30 cm and 150 ms of the temporal coordinate. The grid is "virtual" in the sense that we use a hashmap instead of an actual grid to keep the memory consumption low.

With this grid, we can use a simple flood fill to find all points belonging to the same trajectory. Starting at a non-empty grid cell, we "grow" into neighbouring non-empty cells. These connected components of the 4D grid contain the points of our trajectories. Each component is a single trajectory.

The resolution of the grid is chosen to prevent accidental merging of neighbouring trajectories.

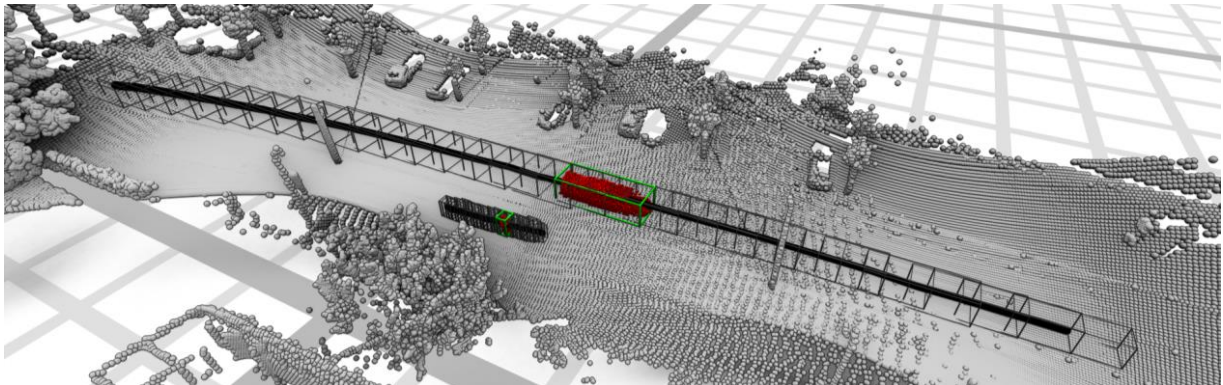

**Figure 5 - After the background separation and global registration, a volumetric space-time flood fill is used to extract points belonging to the same trajectory.**

*Trajectory optimization*

After finding all points belonging to the same trajectory, we can improve their quality. We have a set of 3D points with timestamps belonging to different LiDARs. These points are samples on the surface of the traffic participant. Note that due to the way the LiDARs work, only small subsets of the points (less than 64) are simultaneous in time. Each LiDAR only has a partial view of the vehicle or pedestrian and the next LiDAR might only sample the surface 50-100 ms later. For fast-moving vehicles, this can easily amount to 1 m offset or more. Thus, frame by frame methods are difficult to apply as "frames" would be distorted and only cover the surface partially. Instead, we employ a global registration-like method that alternates between optimizing the trajectory, given a local-space point cloud of the vehicle, and optimizing the local-space point cloud, given a trajectory. The result is a point cloud model of the traffic participant and trajectory data in the form of timestamped positions of the model centre of gravity.

*Anonymization and data privacy*

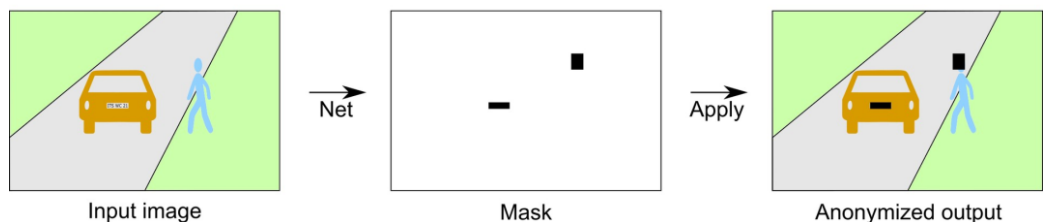

**Figure 6 - Schematic of the anonymisation process: A binary mask obscuring any personal information is created and multiplied with the original image.**

To preserve anonymity, any personal information like faces and license plates needs to be obscured. This implies several requirements: Robustness to changing conditions like weather, sufficient processing speed, and accuracy. Neural networks perform well in all of these areas, hence we suggest utilizing them in our real-time anonymization procedure:

The network generates a mask that is put over the image to conceal any privacy relevant data, see Fig. 6. In order to do that, we leverage a U-Net-like convolutional architecture [10]:

For a single input window, we perform an encoding to a much smaller representation using a fully connected network with ReLu and max pooling for feature extraction. We then inflate the input window to its original size again through fully connected layers and skip connections. A simple squared error loss between predicted- and ground-truth mask is used, re-weighted to account for the sparsity of masked pixels.

This network is then applied in a 2D-convolutional fashion over the whole input image. By taking information from neighbouring windows in consideration for the decoder, the generated masks become less noisy and thus more robust in their output shape.

The network is trained on various datasets for faces and license plates. Training on varying numbers of occurring objects to be masked is not necessary, as the convolutional architecture handles these cases implicitly. This is an important advantage over architectures like YOLO [11] that can only identify a limited number of objects defined by the architecture.

**Data communication**

*Radio relay*

The 5 GHz directional radio range, which is in the frequency band of the public Wi-Fi, must have a minimum distance between the point-to-point data transmission units of 100 m. This alone is a suboptimal prerequisite for a test demonstrator in the municipal area in order to be able to reliably transmit mass data from several data sources in the area of a traffic junction, for example. This requirement alone is a suboptimal prerequisite for a test demonstrator in the municipal area in order to be able to reliably transmit mass data from several data sources in the area of a traffic junction, for example.

Besides, it must be taken into account that in parallel operation of two or more directional radio links in the 5 GHz range, interference-free data transmission cannot be guaranteed due to the unavoidable collision of the Fresnel zones of the directional radio corridors. This means that 5 GHz radio relay technology does not appear to be suitable for use in inner-city areas.

For applications outside the city centre, data transmission in the public 5 GHz or 60 GHz WLAN network based on directional radio can be an alternative that must be verified on a case-by-case basis. In particular, it must be verified to what extent the interference from the public use of the Wi-Fi allows sufficiently high data rates for the transmission of mass data.

*Wi-Fi*

Wi-Fi has undergone a long evolution over the last 20 years. The performance, speed and range have been constantly improved. Responsible for this is the Institute of Electrical and Electronics Engineers (IEEE) in New York.

In 2013, the current Wi-Fi standard 802.11ac was released. Wi-Fi 802.11ac only works in the 5 GHz band, which excludes backward compatibility with 802.11b and 802.11g.

Wi-Fi 802.11ax, or Wi-Fi 6 as it is referred to, is the latest Wi-Fi standard, offering a host of improvements in speed, security and connectivity, and is designed to enable a wireless future with a wide range of devices.

Technically, Wi-Fi 6 will have a 37 % faster single-user data rate (gross data rate 9,600 Mbps) than 802.11ac in perfect connectivity situations by improving data encoding, resulting in higher throughput. Also, the data connection is made more efficient.

The Wi-Fi 6 (Wi-Fi-ax) standard uses both the 2.4 and 5 GHz bands and enables a much higher speed than Wi-Fi-ac. In addition, a Wi-Fi 6 router transmits on a large number of Wi-Fi channels simultaneously and can increase throughput. With the "Spatial Reuse" function, routers make smarter decisions when timing data transmissions so as not to interfere with each other.

From 2021, the 6 GHz frequency band will also be released under the marketing term WiFi 6E, which is intended for short distances. It is a new standard that builds on Wi-Fi 6 and uses the 6 GHz frequency

Highly accurate digital traffic recording as a basis for future mobility research: Methods and concepts of the research project HDV-Mess band for Wi-Fi connections for the first time.

For traffic applications, this performance improvement can provide the basis for the transmission of traffic-related mass data in heavily frequented municipal areas. Within the framework of the research demonstrator, communication is to take place via Wi-Fi, and the transmission of consolidated result data from the LiDAR sensors in connection with the camera signals from the detection locations is to be realised and tested.

*5G*

5G stands for the fifth generation of the standard for mobile telecommunications and mobile internet. The fifth generation of mobile communications is to achieve data rates of up to 10 gigabits per second and is to ensure real-time communication for humans and machines. 5G technology is also about networking machines that communicate in the so-called Internet of Things (IoT). In terms of machine-to-machine communication, this is intended to place more emphasis on energy efficiency. With 5G, the functions of the core network are virtualised via a cloud infrastructure, for example. 5G technology makes the network far more flexible and faster, as several networks can be operated virtually in parallel with the same number of computers.

The 5G mobile network will provide a significant boost to digitalisation worldwide. Machine-to-machine communication will also be considerably facilitated and improved. With the help of new mechanisms, the security of the 5G network will be significantly improved compared to its predecessors.

**Data validation**

For the validation of data collected by mobile ITS-Ss, the comparison between their extracted trajectories and ground truth data providing the exact positions of all perceived road users would be desirable. In contrast to simulations, this is not possible for a setup in the real world. This is why a second method is needed to collect reference data at higher accuracy and precision in comparison to the ITS-Ss, which are referred to as the system under test (SUT) in this section. A well-established method, as shown in [1], that we aim to use for HDV-Mess is to steer a test vehicle equipped with numerous sensor systems through the scenes recorded by our mobile setup, serving two purposes:

First, this vehicle records its trajectory using a state-of-the-art global navigation satellite system (GNSS)-receiver combined with both real-time kinematics (RTK) and an inertial measurement unit (IMU) which have shown to achieve a final positional accuracy of 2 cm [12]. In an offline step, the test vehicle is identified in the SUT's recordings and the corresponding trajectories are compared. Second, the test vehicle uses multiple environment sensors to also perceive all other road users in its FOV. With the same methodology as described in the previous sections, objects are detected in the sensors' raw data, fused, and tracked to obtain precise trajectories. For being able to use this data as a reference, the sensors mounted to the vehicle must outperform the ones on the ITS-Ss in terms of spatial and temporal resolution, for which the lower perspective and thus shorter distance to the measured objects is advantageous. The first approach using a GNSS delivers reference data at a minimum level of uncertainty for a reasonable effort. Another advantage is that the test vehicle together with its IMU may operate under all possible environmental conditions meaning that it can be used to validate our setup at night or in the rain. In contrast, only the SUT's perception of a single vehicle can be evaluated making this method less effective than the second one, with which a larger number of road users of all types are captured. Its drawbacks are, though, the lower accuracy as errors in the ego-movement and the perception of the test vehicle add up, as well as possible sensor problems in difficult conditions.

Finally, even the test vehicle's environment sensors cannot perceive all objects in the scene, which is why we employ another additional validation method. [13] shows that a static drone flying over an SUT can also serve as a reference sensor due to its high accuracy and the very low effort needed to operate it. Since this approach only functions in clear weather conditions and is subject to legal flight restrictions, it cannot replace the first two methods but is useful to complement them for comprehensive validation.

**Conclusion and outlook**

In our work, we have shown that the use of mobile, modular, and fully self-sufficient ITS-Ss offers great potential for the individual collection of traffic data sets. The mobile devices, which are designed in the Highly accurate digital traffic recording as a basis for future mobility research: Methods and concepts of the research project HDV-Mess form of trailers, allow flexible deployment at a wide variety of road cross-sections in urban and rural areas, as well as on highways, in coordination with local authorities. Due to the modularity, different sensors for traffic detection can be tested and evaluated with regard to their suitability for the respective application. The innovative mobile and modular ITS-Ss are rounded off by full self-sufficiency, which enables continuous measurements in real traffic for several weeks. By using state-of-the-art algorithms, all collected raw sensor data is first anonymized and subsequently processed into highly accurate trajectories. This creates a valuable data basis for naturalistic traffic data for research purposes of automated driving.

Now that the conceptual design and development of our mobile and modular ITS-Ss has been completed, we intend to collect traffic data at different measurement cross-sections in the further course of the project. Finally, we will analyse the collected data and evaluate our overall concept for highly accurate traffic data collection with the help of reference measurements.

**Acknowledgement**

The research leading to these results is funded by the European Regional Development Fund (ERDF) within the project "HDV-Mess: Highly accurate digital traffic recording as a basis for future mobility research - Construction of mobile and modular measuring stations" (EFRE-0500038). The authors would like to thank the consortium for successful cooperation.

**References**

1. Sun, P., Kretzschmar, H., Dotiwalla, X., Chouard, A., Patnaik, V., Tsui, P., ... & Vasudevan, V. (2020). Scalability in perception for autonomous driving: Waymo open dataset. In *Proceedings of the IEEE/CVF Conference on Computer Vision and Pattern Recognition* (pp. 2446-2454).

2. Krajewski, R., Bock, J., Kloeker, L., & Eckstein, L. (2018, November). The highd dataset: A drone dataset of naturalistic vehicle trajectories on german highways for validation of highly automated driving systems. In *2018 21st International Conference on Intelligent Transportation Systems (ITSC)* (pp. 2118-2125). IEEE.

3. Meissner, D., & Dietmayer, K. (2010, June). Simulation and calibration of infrastructure based laser scanner networks at intersections. In *2010 IEEE Intelligent Vehicles Symposium* (pp. 670-675). IEEE.

4. Schnieder, L., & Lemmer, K. (2012). Anwendungsplattform Intelligente Mobilität-eine Plattform für die verkehrswissenschaftliche Forschung und die Entwicklung intelligenter Mobilitätsdienste. *Internationales Verkehrswesen*, *64*(4), 62-63.

5. Köster, F., Mazzega, J., & Knake-Langhorst, S. (2018). Automatisierte und vernetzte Systeme Effizient erprobt und evaluiert. *ATZextra*, *23*(5), 26-29.

6. Krämmer, A., Schöller, C., Gulati, D., & Knoll, A. (2019). Providentia-a large scale sensing system for the assistance of autonomous vehicles. *arXiv preprint arXiv:1906.06789*.

7. Keiser, J., Schäufele, B., Bunk, S., Berger, M., Völker, S., Wilhelm, M., ... & Heßler, A. (2018). Die digital vernetzte Protokollstrecke-urbanes Testfeld automatisiertes und vernetztes Fahren in Berlin: Schlussbericht der TU Berlin und Fraunhofer-Gesellschaft: DIGINET-PS: Berichtszeitraum: 04/2017-12/2019. *Computer Communications*, *122*, 93-117.

8. Kloeker, L., Kloeker, A., Thomsen, F., Erraji, A. & Eckstein, L. (2020). Traffic Detection Using Modular Infrastructure Sensors as a Data Basis for Highly Automated and Connected Driving. 29. Aachen Colloquium - Sustainable Mobility, 29(2), 1835-1844.

9. Gao W, Tedrake R (2019). FilterReg: Robust and Efficient Probabilistic Point-Set Registration Using Gaussian Filter and Twist Parameterization. In Proceedings of the *IEEE/CVF Conference on Computer Vision and Pattern Recognition (CVPR)*. pp. 11095-11104.

10. Ronneberger, O., Fischer, P. & Brox, T. (2015). U-Net: Convolutional Networks for Biomedical Image Segmentation. In *Medical Image Computing and Computer-Assisted Intervention (MICCAI) 2015*, Springer, LNCS, Vol.9351: pp. 234-241.

11. Redmon, J., Divvala, S., Girshick, R. & Farhadi, A. (2015). You Only Look Once: Unified, Real-Time Object Detection. In *2016 IEEE Conference on Computer Vision and Pattern Recognition (CVPR)*.

12. Brahmi, M., Siedersberger, K. H., Siegel, A., & Maurer, M. (2013). Reference Systems for Environmental Perception: Requirements, Validation and Metric-based Evaluation. In *6. Tagung Fahrerassistenzsysteme*.

13. Krajewski, R., Bock, J., Eckstein, L. (2019). Drones as a Tool for the Development and Safety Validation of Highly Automated Driving. In Proceedings *28th Aachen Colloquium Automobile and Engine Technology 2019*. Aachen, pp. 1449-1461.